\documentclass[10pt,twocolumn,letterpaper]{article}
\PassOptionsToPackage{table,dvipsnames}{xcolor}
\usepackage{float}    
\usepackage{placeins} 

\usepackage{hyperref} 
 \usepackage{booktabs}
\usepackage[numbers,sort&compress]{natbib}
\makeatletter
\let\ESO@isMEMOIR\relax
\makeatother
\usepackage{ifpdf}
\ifpdf
    \usepackage{pdfpages}
\fi

\usepackage{amssymb}

\definecolor{iccvblue}{rgb}{0.21,0.49,0.74}

\def\paperID{15670} 
\def\confName{ICCV}
\def\confYear{2025}

\title{Dual Codebook VQ: Enhanced Image Reconstruction with Reduced Codebook Size}

\author{
Parisa Boodaghi Malidarreh, Jillur Rahman Saurav, Thuong Le Hoai Pham, Amir Hajighasemi,\\
Anahita Samadi, Saurabh Shrinivas Maydeo, Mohammad Sadegh Nasr, Jacob M. Luber\\\\
Department of Computer Science, The University of Texas at Arlington\\
{\tt jacob.luber@uta.edu}
}
\usepackage{soul} 
\usepackage{amsmath}  
\usepackage{graphicx} 
\usepackage{multirow}
\usepackage[table]{xcolor}
\date{}
\begin{document}
\maketitle

\begin{abstract}
Vector Quantization (VQ) techniques face significant challenges in codebook utilization, limiting reconstruction fidelity in image modeling. We introduce a Dual Codebook mechanism that effectively addresses this limitation by partitioning the representation into complementary global and local components. The global codebook employs a lightweight transformer for concurrent updates of all code vectors, while the local codebook maintains precise feature representation through deterministic selection. This complementary approach is trained from scratch without requiring pre-trained knowledge. Experimental evaluation across multiple standard benchmark datasets demonstrates state-of-the-art reconstruction quality while using a compact codebook of size 512 - half the size of previous methods that require pre-training. Our approach achieves significant FID improvements across diverse image domains, particularly excelling in scene and face reconstruction tasks. These results establish Dual Codebook VQ as an efficient paradigm for high-fidelity image reconstruction with significantly reduced computational requirements.
\end{abstract}    
\section{Introduction}
\label{sec:intro}

Vector Quantization (VQ) \cite{gray1984vector} has emerged as a core technique for mapping high-dimensional continuous data—such as images or feature representations—into a discrete latent space defined by a finite set of learned code vectors. By replacing input features with the nearest codebook entries, VQ enforces a discrete bottleneck that enables more efficient and compact representations. This approach has been instrumental in many unsupervised learning tasks, notably in image generation, as demonstrated by Vector Quantized Variational Autoencoders (VQ-VAE) and related methods \cite{esser2021taming, van2017neural,razavi2019generating}. These successes underscore the importance of VQ in bridging discrete and continuous modeling paradigms for modern generative models.
\begin{figure}[!ht]
    \centering
    \includegraphics[width=\columnwidth]{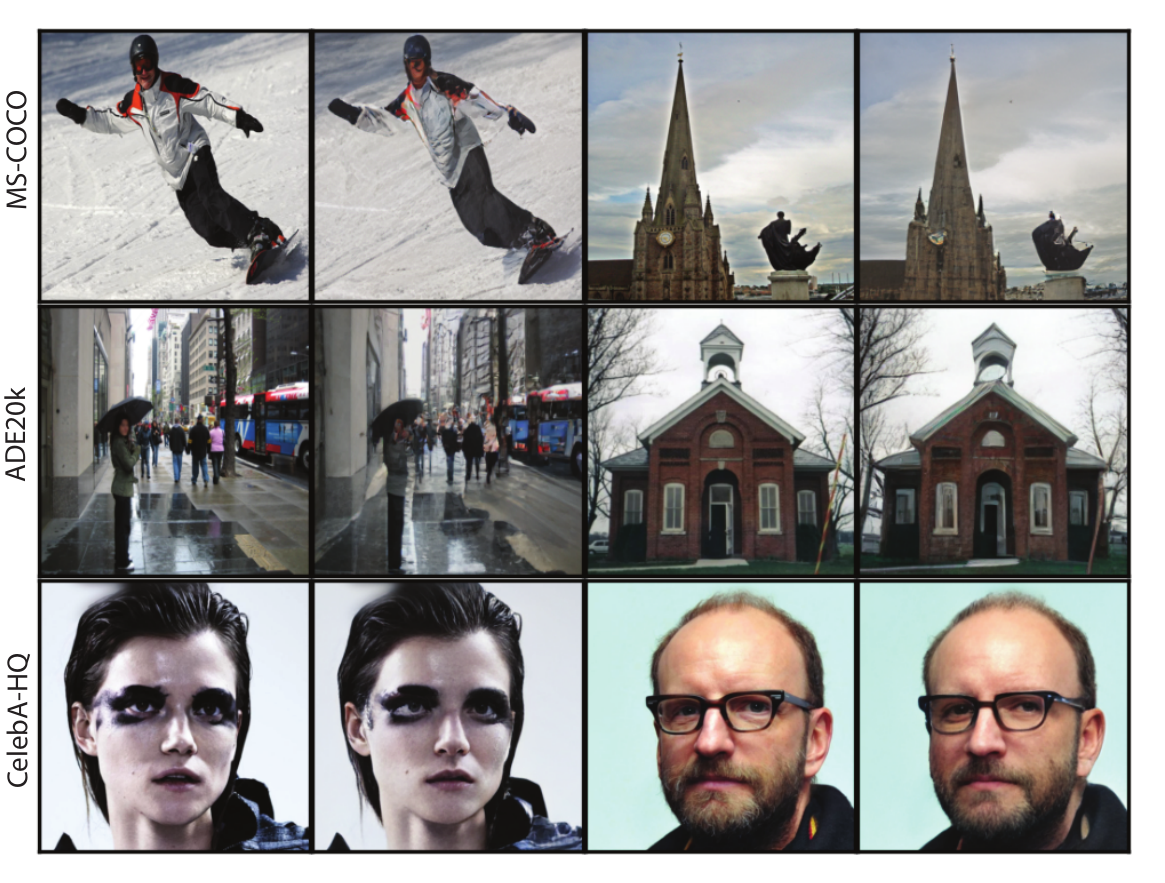}
    \caption{Comparison of reconstruction quality across different datasets. Our Dual Codebook method produces sharp details and  preserves textures across different domain}
    \label{fig:example-reconstructions}
\end{figure}
Despite its effectiveness, deterministic quantization can lead to degraded image quality, especially if the codebook is poorly structured in the early stages of training. A common issue is codebook collapse, where only a small subset of code vectors receive significant updates while the rest remain underutilized \cite{takida2022sq}. Previous research has introduced various modifications—such as adversarial loss \cite{esser2021taming}, stochastic quantization, and refined codebook update rules \cite{zhang2023regularized, zheng2023online, zhang2024codebook, zhu2024scaling, zhu2024addressing, guotao2025lg}—to improve image fidelity and codebook usage. However, many of the highest-performing approaches rely on pre-trained models (e.g., codebook priors or alignment modules) \cite{zhang2024codebook, guotao2025lg, chen2024softvq}, arguing that starting from scratch often results in suboptimal codebooks.

In this paper, we propose a Dual Codebook \footnote{Code available at: \href{https://github.com/jacobluber/DualCodebookVQ/}{https://github.com/jacobluber/DualCodebookVQ/}} mechanism that tackles these challenges without relying on pre-trained knowledge. Inspired by recent advances in non-deterministic codebook optimization \cite{zhang2024codebook, zheng2023online}, our method fuses deterministic and stochastic elements to refine image quality and maximize codebook utilization. 

Specifically, we split the latent representation into two separate pathways: one dedicated to local feature updates using deterministic selection, and another for global feature optimization via a lightweight transformer that updates all code vectors simultaneously. This dual strategy retains the advantages of conventional VQ while mitigating codebook collapse and reducing reliance on large pre-trained distributions.
Our main contribution is that we propose a Dual Codebook approach that effectively captures both global and local features from input data without increasing the codebook size or decreasing throughput.
Our method is designed to be trained from scratch without incorporating any prior knowledge, making it perform well on diverse the datasets we tested on and eliminating the need for domain-specific information across different datasets.
Our Dual Codebook demonstrates superior performance compared to previous state-of-the-art models on most datasets, achieving these results with a smaller codebook size. Notably, we achieve significant FID improvements on ADE20K (17.03 vs. 20.25) and MS-COCO (4.19 vs. 9.82) compared to VQCT, despite using a codebook less than one-tenth the size.

\begin{figure*}
    \centering
    \includegraphics[width=0.9\textwidth]{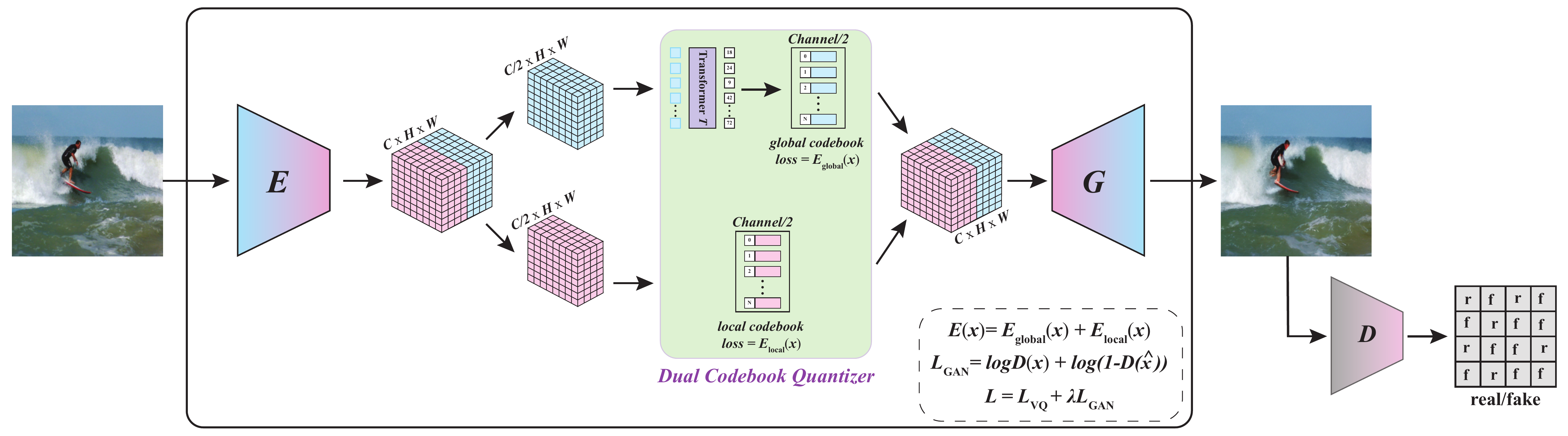}
    \caption{Overview of our Dual Codebook framework, comprising an encoder, a dual codebook mechanism, a decoder, and a discriminator. The encoder processes an input image into spatially continuous vectors, which are then split into two halves. The first half is processed using a lightweight transformer called global codebook, while the second half is updated via a deterministic quantizer as a local codebook. After mapping each continuous vector to discrete code vectors, the two halves are concatenated and fed into the decoder to reconstruct the image using the quantized representations. A discriminator is employed to incorporate GAN-based training objectives.} 
    \label{fig:overview-architecture}
\end{figure*}

Empirical results demonstrate that our approach produces superior reconstructions (Figure~\ref{fig:example-reconstructions}) while being both computationally efficient and easy to integrate into standard VQ frameworks like VQ-GAN, as illustrated in Figure~\ref{fig:overview-architecture}.

\section{Related Work}

In order to effectively constructing well-structured codebooks for latent representation, a key challenge in VQ-based models is designing an efficient codebook update method. Improper updates can lead to an imbalanced utilization of code vectors, potentially limiting representation quality and affecting overall model performance. Multiple solutions for robust codebook learning have been proposed, including codebook update~\cite{dhariwal2020jukebox, razavi2019generating, williams2020hierarchical, zheng2023online}, quantization strategies~\cite{baevski2019vq, takida2022sq, vuong2023vector}, or regularization~\cite{zhang2023regularized}. For example, SQ-VAE~\cite{takida2022sq} improves the VQ-VAE by introducing stochastic quantization and a learnable categorical posterior, enhancing flexibility in mapping inputs to discrete codes. VQ-WAE~\cite{vuong2023vector} builds on SQ-VAE by encouraging uniform usage of codebook entries through Wasserstein distance regularization, improving codebook utilization. Jukebox~\cite{dhariwal2020jukebox} address underutilized codevectors with codebook resets, and randomly reinitializing inactive entries. CVQ-VAE~\cite{zheng2023online} handles feature drift with online clustering on top of Jukebox-style codebook resets. Reg-VQ~\cite{zhang2023regularized} stabilizes codebook usage and balances reconstruction with probabilistic regularization.

The current state-of-the-art model in vector-quantized image modeling is VQCT, which leverages a pretrained codebook and part-of-speech knowledge as priors, along with a graph convolutional codebook transfer network to generate the codebook. Despite its strong performance it relies on a pretrained codebook from language model with codebook of size 6207 (1949 adjectives and 4258 nouns) which is a large codebook size ~\cite{zhang2024codebook}. In contrast, our work addresses these limitations by: (1) introducing a dual-codebook architecture integrated with a transformer, which achieves higher performance across multiple evaluation datasets with a smaller codebook size while significantly improving codebook utilization; and (2) designing a framework that does not rely on any pre-trained multimodal knowledge or domain-specific information, yet still surpasses previous state-of-the-art approaches and demonstrates superior performance in image synthesis across various datasets.

\section{Method}
\label{sec:method}
\subsection{Background: VQ-VAE and VQ-GAN}

Vector Quantization (VQ) enables the discrete representation of continuous data through a learned codebook and has been widely applied in image modeling. Oord et al.'s VQ-VAE~\cite{van2017neural} first introduced this idea by replacing the prior of a Variational Autoencoder with a discrete codebook, inspiring many extensions to improve reconstruction quality and codebook expressiveness. VQ-VAE establishes a discrete latent space using vector quantization. It encodes an image \( x \in \mathbb{R}^{H \times W \times c} \) into a discrete latent representation. The encoder maps \( x \) to a continuous latent space \( Z \in \mathbb{R}^{h \times w \times n_c} \), which is then quantized by replacing each \( \hat{z}_i \) with the nearest codebook vector \( e_k \). The decoder reconstructs \( \hat{x} \) from \( z_q \). The training objective consists of three components: a reconstruction loss, a codebook loss, and a commitment loss weighted by \( \beta \), ensuring encoder stability.  

Many works extend these core ideas. In \cite{razavi2019generating}, a hierarchical model is introduced, where latent variables are structured at multiple scales, each utilizing a similar quantization process to capture both global and local structures. In \cite{baevski2019vq}, the discrete representation framework is applied to audio, where continuous speech features \( z_e(x) \) are quantized into a finite set \( \{ e_j \} \) via a nearest-neighbor search, enabling efficient self-supervised learning. Among these extensions, VQ-GAN \cite{esser2021taming} integrates VQ-VAE’s discrete latent space with adversarial training to improve perceptual quality. VQ-GAN introduces a GAN-based loss along with a perceptual loss computed over high-level features, leading to sharper and more realistic reconstructions. These advancements highlight the broad applicability of vector quantization in generative modeling. 




\subsection{Background: Transformers}
Transformers~\cite{vaswani2017attention} are renowned for their ability to model long-range dependencies through self-attention mechanisms, enabling the efficient processing and understanding of sequential data. Since their introduction, this architecture has been rapidly adopted across a wide range of modalities, including text, images, and signals, consistently demonstrating a remarkable capacity to capture complex relationships and uncover the underlying structure of information.

In the context of latent-space modeling, particularly within frameworks like VQ-VAE~\cite{van2017neural}, managing and enhancing codebook utilization remains a central challenge. Regularizing, updating, and transforming codebook vectors have emerged as effective strategies to address this issue~\cite{dhariwal2020jukebox, razavi2019generating, williams2020hierarchical, zheng2023online, baevski2019vq, takida2022sq, vuong2023vector, zhang2023regularized}. In this regard, we believe that Transformers are particularly well-suited for operating on discrete latent representations, offering a principled way to maximize codebook efficiency and expressiveness. By leveraging self-attention, Transformers can model intricate dependencies between latent tokens, which is essential for capturing both global context and local coherence within the compressed latent space. This holistic view over the entire latent sequence enables Transformers to iteratively refine and enrich latent representations, allowing each token embedding to absorb increasingly complex contextual information, effectively blending fine-grained details with high-level structural cues.

\begin{table*}
  \centering
  \caption{Results of image reconstruction on ADE20K, CelebA-HQ, and MS-COCO. The best results are highlighted in bold. The reported results were obtained by tracking the reconstruction loss on the validation set, selecting the best checkpoint, and evaluating it on the test split.}
  \label{tab:results}
  \resizebox{\textwidth}{!}{ 
  \begin{tabular}{l c cccc cccc cccc }
    \toprule
    \multirow{2}{*}{Models} & \multirow{2}{*}{Codebook Size} & \multicolumn{4}{c}{ADE20K~\cite{zhou2017scene,zhou2019semantic}} & \multicolumn{4}{c}{CelebA-HQ~\cite{liu2015deep}} & \multicolumn{4}{c}{MS-COCO~\cite{lin2014microsoft}} \\
    \cmidrule(lr){3-6} \cmidrule(lr){7-10} \cmidrule(lr){11-14} 
    &  & FID$\downarrow$ & PSNR$\uparrow$ & $L_1\downarrow$ & $L_2\downarrow$ 
    & FID$\downarrow$ & PSNR$\uparrow$ & $L_1\downarrow$ & $L_2\downarrow$ 
    & FID$\downarrow$ & PSNR$\uparrow$ & $L_1\downarrow$ & $L_2\downarrow$ \\
    \midrule
    VQ-VAE~\cite{van2017neural} & 1024 & 116.85 & 21.08 & 0.1282 & 0.0368  
    & 36.08  & 25.29 & 0.0719 & 0.0139
    & 86.21  & \textbf{23.55} & 0.0933 & 0.0226 \\
    VQ-GAN ~\cite{esser2021taming} & 1024 & 22.04  & 20.42  & 0.1290 & 0.0451  
    & 5.66   & 24.10  & 0.0798 & 0.0175  
    & 14.45  & 20.21  & 0.1311 & 0.0475 \\
    Gumbel-VQ~\cite{baevski2019vq} & 1024 & 24.12 & 20.04 & 0.1359 & 0.0482  
    & 6.22  & 23.65 & 0.0837 & 0.0194  
    & 15.30 & 20.00 & 0.1354 & 0.0488 \\
    CVQ~\cite{zheng2023online} & 1024 & 33.63 & 19.91 & 0.1379 & 0.0486  
    & 5.19  & 23.15 & 0.0917 & 0.0214  
    & 9.94  & 20.48 & 0.1253 & 0.0443 \\
    
    VQCT~\cite{zhang2024codebook} & 6207 & 20.25 & \textbf{21.30} & 0.1144 & 0.0374  
    & \textbf{5.02}  & 25.18 & 0.0699 & 0.0134  
    & 9.82  & 21.46 & 0.1108 & 0.0366 \\
    
   \cellcolor{gray!20} \textbf{Dual Codebook (Ours)}  & \cellcolor{gray!20}\textbf{512} & \cellcolor{gray!20}\textbf{17.03} & \cellcolor{gray!20} 20.88 & \cellcolor{gray!20}\textbf{0.0682} & \cellcolor{gray!20}\textbf{0.0099}  
    & \cellcolor{gray!20} 8.61  & \cellcolor{gray!20}\textbf{25.88} & \cellcolor{gray!20}\textbf{0.0364} & \cellcolor{gray!20}\textbf{0.003}  
    & \cellcolor{gray!20}\textbf{4.19}  &  \cellcolor{gray!20}20.72 & \cellcolor{gray!20}\textbf{0.0717} & \cellcolor{gray!20}\textbf{0.0103} \\
    
    \bottomrule
  \end{tabular}
  }
\end{table*}

\subsection{Dual Codebook}
Many studies have focused on enhancing reconstructed image quality by optimizing codebook utilization and mitigating codebook collapse. Most existing approaches introduce stochastic updates to the codebook, which have proven effective in improving performance. In this work, we leverage the advantages of both deterministic and stochastic update strategies for codebook learning from scratch, without incorporating any prior knowledge or predefined distributions. To achieve this goal, we propose a novel Dual Codebook Mechanism, which employs two distinct update methods while maintaining the original codebook size and code vector dimensionality. The key objective is to eliminate the need for pre-processing to establish prior distributions, avoid increasing model complexity, and maximize the benefits of existing codebook learning techniques.

As illustrated in Figure~\ref{fig:overview-architecture}, our proposed Dual Codebook Mechanism captures both global and local image features through a Dual Codebook optimization strategy, which are labelled as the global and local codebooks. This approach ensures the reconstruction of high-quality images while maximizing the utilization of all code vectors. The overall architecture follows a VQGAN-like structure, consisting of four main components: an Encoder, a Decoder, a Vector Quantization (VQ) module and Discriminator. The encoder-decoder architecture in our architecture compresses 256x256 RGB images into a discrete latent space using a hierarchical convolutional encoder with 16× downsampling. The discriminator, a PatchGAN-style CNN, operates on RGB images, starting after 10,000 steps to stabilize training, with a weight of 0.8 in the loss function. The model is trained with learning rate of 4.5e-6 to ensure stable optimization.

However, unlike standard VQGAN, the key distinction in our method lies in the vector quantization stage, where we employ two separate codebooks of equal size, referred to as the global codebook and the local codebook. Given an input image of size $(H, W)$, the encoder transforms it into a continuous representation of shape $(Channel, H, W)$. This representation is then split into two separate tensors of shape $(Channel/2, H, W)$. The first half is processed using a lightweight transformer-based update mechanism, enhancing the codebook with learned contextual information. The transformer we used for global codebook consists of a 6-layer Transformer Encoder, where each layer utilizes an embedding dimension of half the codebook dimensionality, 8 attention heads, a feedforward dimension of 2048. This architecture is designed to efficiently update the codebook while maintaining computational efficiency. The second half undergoes a deterministic update, where the nearest code vector is selected in each iteration. The two updated codebooks are then concatenated to form the final quantized representation of shape $(Channel, H, W)$, which is subsequently fed into the decoder to reconstruct the image. The corresponding loss function is defined in Equation \ref{eq:total_loss} which is similar to VQ-GAN consists of two parts vector quantization loss is shown with $L_{VQ}$ and GAN loss that is shown with $L_{GAN}$. Here, \( E \) refers to the encoder, \( G \) to the decoder, \( Z \) to the quantizer, and \( D \) to the discriminator.

\begin{equation}
\resizebox{\columnwidth}{!}{$
    Q^* = \arg \min_{E,G,Z} \max_D \mathbb{E}_{x \sim p(x)} \Big[
    L_{\text{VQ}}(E, G, Z) + \lambda L_{\text{GAN}}(\{E, G, Z\}, D) \Big]
$}
\label{eq:total_loss}
\end{equation}

Similar to VQ-GAN, the adaptive weight $\lambda$ should be computed according to equation \ref{eq:lambda}, where $\nabla_{G _L} [\cdot]$ denotes the gradient of its input and $\delta = 10^{-6}$ is used for numerical stability.

\begin{equation}
\lambda = \frac{\nabla_{G_L} [\mathcal{L}_{\text{rec}}]}{\nabla_{G_L} [\mathcal{L}_{\text{GAN}}] + \delta}
\label{eq:lambda}
\end{equation}

Similar to vanilla VQ-GAN we have $L_{VQ}$ which can be seen in Equation \ref{eq:vqloss}.
\begin{equation}
\resizebox{\columnwidth}{!}{$
L_{\text{VQ}}(E, G, Z) = \mathcal{L}_{\text{rec}} + \|\text{sg}[E(x)] - z_q\|_2^2 + \|\text{sg}[z_q] - E(x)\|_2^2
$}
\label{eq:vqloss}
\end{equation}

The main difference between normal VQ-GAN and \textbf{Dual Codebook VQ} method lies in the codebook loss part of $L_{VQ}$. The global and local coodebooks comprised in \textbf{Dual Codebook VQ} are defined as: 

\begin{equation}
E(x) = E_{global}(x) + E_{local}(X)
\label{eq:vqloss-codebook}
\end{equation}


 \begin{figure*}[!ht]
    \centering
    \includegraphics[width=\linewidth]{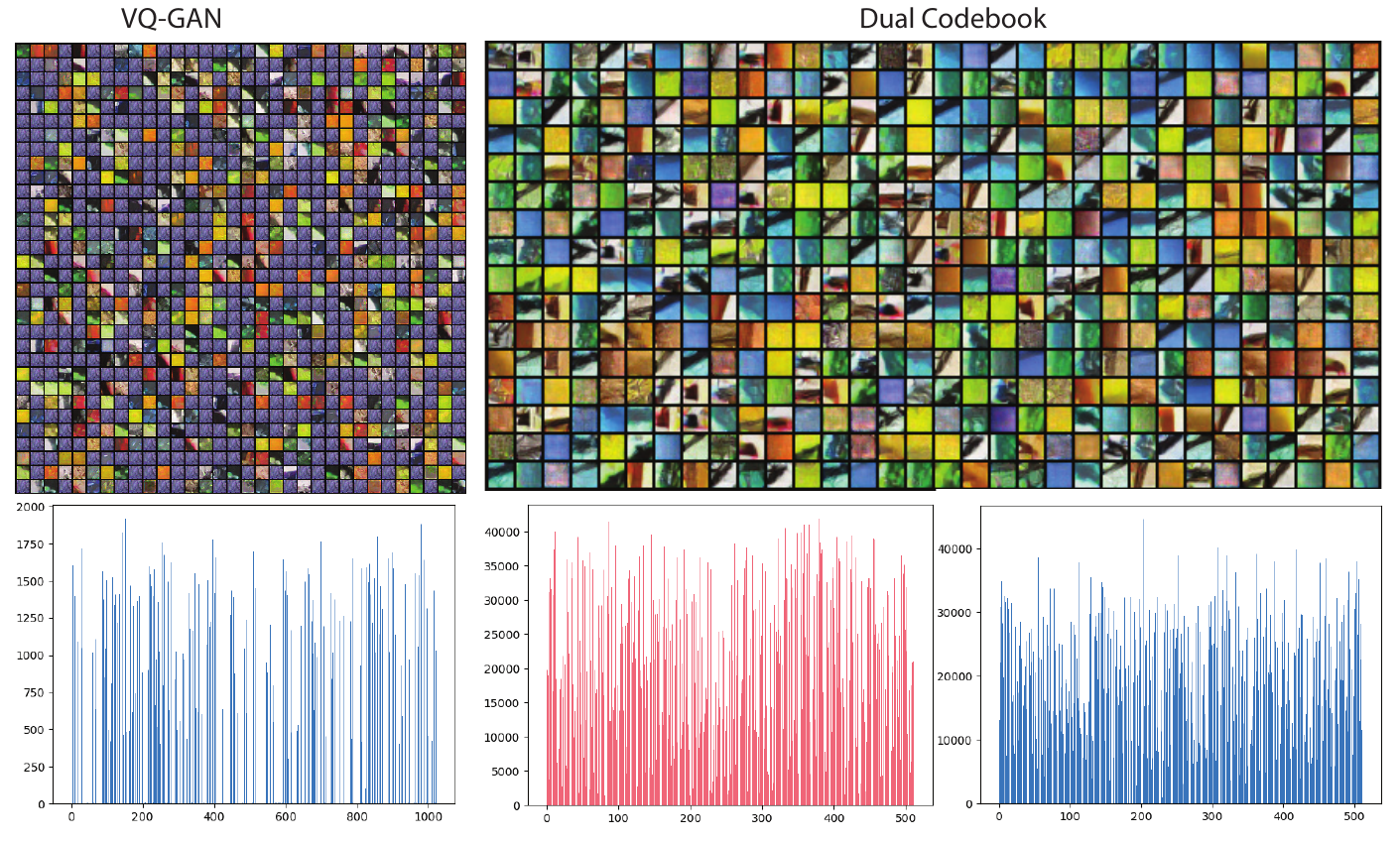}
    \caption{Visualization of codebook (first row) and illustration of codebook usage (second row) on MS-COCO dataset for VQ-GAN (blue) and \textbf{Dual Codebook VQ} (red and blue). In this comparison, VQ-GAN was tested with codebook size of 1024, but our Dual Codebook was teste with 512 codebook size totally (the red graph related to global codebook usage and blue one is for local codebook usage for Dual codebook).}
    \label{fig:utility}
\end{figure*}

\section{Experiments}
\label{sec:experiment}
\subsection{Experimental Details}

\subsubsection{Implementation} The Dual Codebook approach is implemented by modifying the vector quantization function and making slight adjustments to the loss function in PyTorch. Our implementation is based on the VQ-GAN architecture with these targeted modifications. The model is developed with PyTorch Lightning API. All
experiments were conducted on an NVIDIA DGX A100 with 8, 80 GB A100 GPUs.

\subsubsection{Dataset} We assess our method on three publicly available datasets: ADE20K \cite{zhou2017scene,zhou2019semantic}, CelebA-HQ \cite{liu2015deep}, and MSCOCO \cite{lin2014microsoft}. Consistent with the approach in \cite{zhang2024codebook} all images are resized to 256 × 256 for experimental evaluation and the latent size is 16×16.

\subsubsection{Evaluation Metrics} We compared our method with five recent VQ-based models VQ-VAE \cite{van2017neural}, VQ-GAN \cite{esser2021taming}, Gumbel-VQ \cite{baevski2019vq}, CVQ \cite{zheng2023online}, and VQCT \cite{zhang2024codebook} and evaluated them with image reconstruction performance. Four evalution metrics is used including Frechet Inception Score (FID) \cite{heusel2017gans}, $l1$ , $l2$, and PSNR (Peak Signal to noise Ratio) to show the quality of reconstructed images.

\subsection{Main Results}
\subsubsection{Quantitative Evaluation}
We showed the main result of our evaluation in Table \ref{tab:results} where we compare the performance of our reconstructed images against five baseline models under the same compression ratio (\( 768\times \), \textit{i.e.}, \( 256 \times 256 \times 3 \rightarrow 16 \times 16 \)). As shown in the table, our Dual Codebook method outperforms state-of-the-art approaches across most evaluation metrics. Notably, our model demonstrates a significant performance margin compared to VQCT, on ADE20K \cite{zhou2017scene,zhou2019semantic} (17.03 vs. 20.25) and MS-COCO \cite{lin2014microsoft} (4.19 vs
9.82) datasets, highlighting its effectiveness in VQ-based models.
Despite utilizing a codebook of size 512—substantially smaller than VQCT \cite{zhang2024codebook}, which employs a codebook of size 6207 (1949 adjectives and 4258 nouns)—our method still achieves a notable improvement in FID and loss values. It is important to emphasize that, unlike VQCT, which leverages a pre-trained codebook prior for optimization, our Dual Codebook model is trained from scratch, further demonstrating its robustness and adaptability.

\subsubsection{Qualitative Evaluation}
\begin{figure}
    \centering
    \includegraphics[width=\linewidth]{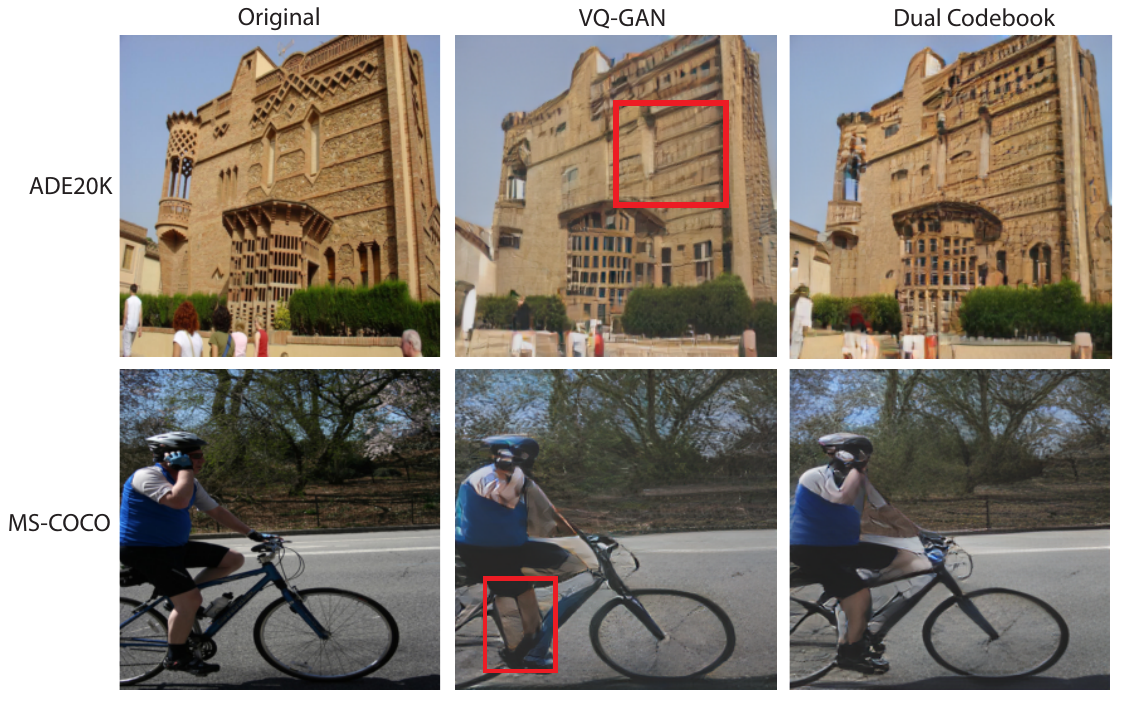}
    \caption{Comparison of reconstructed images from our Dual Codebook and VQ-GAN on two datasets. Two models are trained under the same settings and same compression ration (\( 768\times \), \textit{i.e.}, \( 256 \times 256 \times 3 \rightarrow 16 \times 16 \)) but different codebook size (codebook size for VQ-GAN is 1024 and for Dual Codebook is 512). Our proposed quantization method significantly improves reconstruction quality, particularly enhancing texture details in building facades and the sky background in the ADE20K dataset. In the second row, for MS-COCO, it not only refines the anatomical details of the man but also enhances the reconstruction accuracy and color fidelity of the man's helmet.(the red boxed highlighted the reconstruction details).}
    \label{fig:qualitative-result}
\end{figure}

\begin{figure*}
    \centering
    \includegraphics[width=\textwidth]{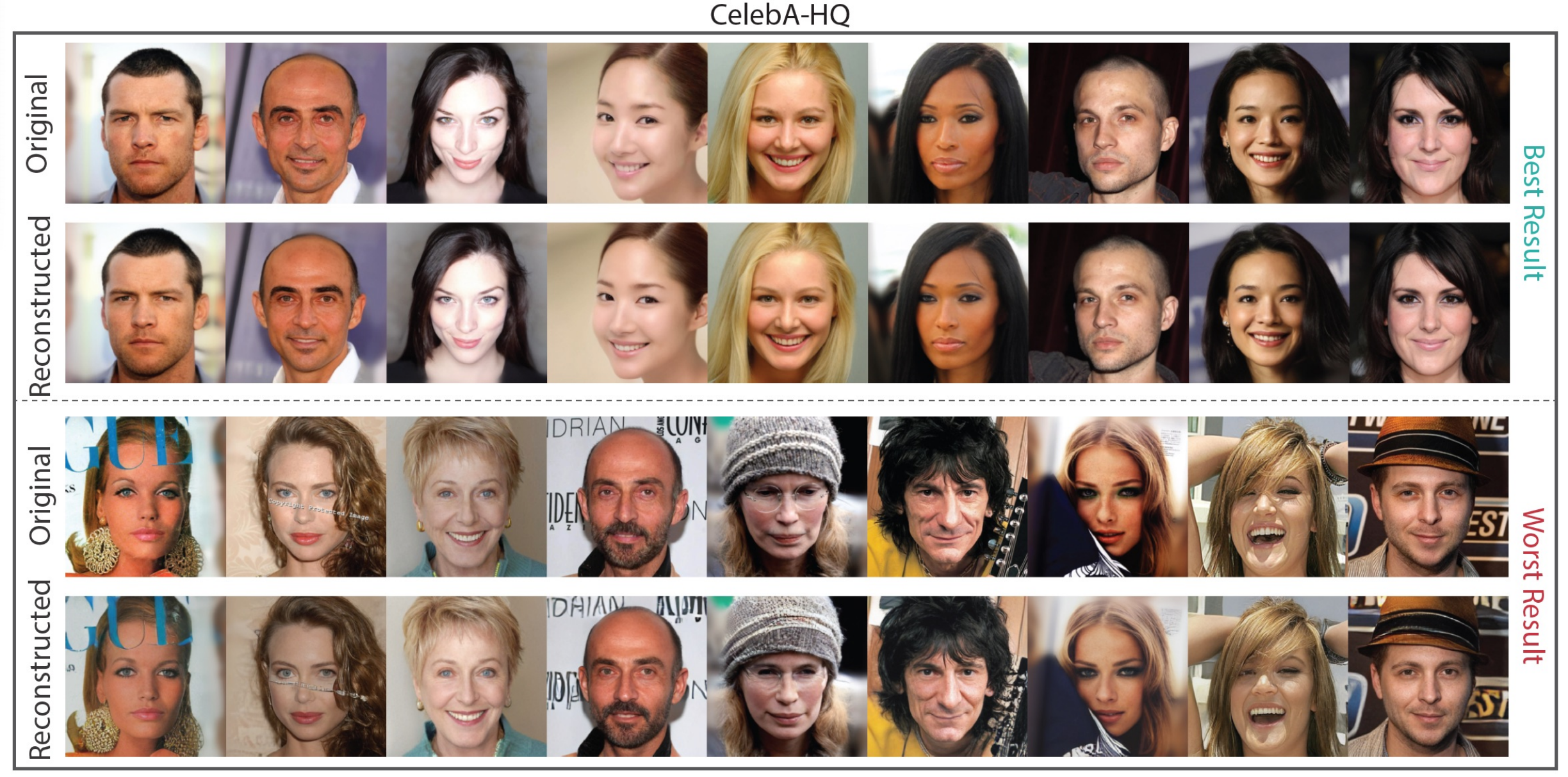}
    \caption{Best and worst reconstruction result on CelebA-HQ (test split) with Dual Codebook quantizer respectively first and second row (Best and worst case selected based on PSNR value).}
    \label{fig:_qualitative-result-cel}
\end{figure*}

The qualitative evaluation of our model is shown in Figure~\ref{fig:qualitative-result} where we compare the reconstruction result of our model with the VQ-GAN as our baseline model on two different dataset ADE20k and MS-COCO. Compared to the baseline VQ-GAN, our model achieves superior reconstruction quality, particularly in preserving textures 
and producing sharper details across different domain. This comparison highlights the effectiveness of our Dual Codebook approach in generating high-fidelity reconstructed images. Additionally, Figure~\ref{fig:_qualitative-result-cel} presents a comprehensive set of reconstruction results on CelebA-HQ dataset using our Dual Codebook method. The figure showcases both the best-case and worst-case reconstructions, providing a detailed qualitative analysis of our model's performance.

\subsection{Ablation Experiments}

\subsubsection{Dual Codebook Partition}
Table \ref{tab:codebook configuration} presents an ablation study evaluating the effectiveness of our Dual Codebook design in enhancing image quality. Specifically, we investigate (1) the impact of introducing global and local codebooks at varying ratios, and (2) the effectiveness of incorporating a transformer into the global codebook.
The experiment compares five configurations: (i) the baseline (VQ-GAN \cite{esser2021taming}); and four Dual Codebook variants using different channel allocations and transformer settings: (ii) 128-channel global and local codebooks without a transformer; (iii) a 192-channel transformer-based global codebook with a 64-channel simple local codebook; (iv) a 64-channel transformer-based global codebook with a 192-channel simple local codebook; and (v) 128-channel transformer-based global and simple local codebooks.

For the ablation study, we chose MS-COCO \cite{lin2014microsoft} as the primary dataset due to its extensive size and complexity, making it the most suitable for scene image reconstruction. As shown in Table~\ref{tab:codebook configuration}, row (i) and (ii) suggested having Dual Codebook yields good performance on all parameters, even without using any non-deterministic updating method like transformer. On the other hand,comparing (ii) and (v) suggested that having transformer-based global codebook resulted in better performance across all metrics. Additionally, having a a large rather than small global codebook yields better results across all evaluation metrics, as shown in comparing (iii) and (v) to (iv). 

This study suggested that implementing Dual Codebook with large transformer-based global codebook size significantly improve model's performance. Between the configurations (iv) and (v), (iv) had slightly better FID value, but (v) presented better losses and PSNR values. Hence, we decided to use configuration (v) to implement the Dual Codebook for all experiments.


\begin{table}
  \centering
  \small
  \caption{Evaluation results of ablation study on MS-COCO dataset with four different codebook configuration. In Global and Local columns we have T for transformer based codebook and S for simple codebook method. In different models we split the 256 channels for global and local as it mentioned in the table. The best results are highlighted in bold. See Supplimental Figure~\ref{fig:supp1-ultility} for detailed codebook utility.}
  \label{tab:codebook configuration}
  \resizebox{\columnwidth}{!}{ 
  \begin{tabular}{l  c ccccc}
    \toprule
    Models & Global & Local  & FID$\downarrow$ & PSNR$\uparrow$ & $L_1\downarrow$ & $L_2\downarrow$ \\
    \midrule

    (i) VQ-GAN~\cite{esser2021taming} & - & - & 14.45 & 20.21 & 0.1311 & 0.0475\\
    (ii) Dual Codebook &  S - 128 & S - 128 & 4.91 & 19.58 & 0.0857 & 0.0131  \\
    (iii) Dual Codebook & T - 192 & S - 64 & 4.67 & 19.26 & 0.0865 & 0.0139\\
    (iv) Dual Codebook & T - 64 & S - 192   & \textbf{4.12} & 19.51 & 0.0853 & 0.0133 \\
    (v) Dual Codebook & T - 128 & S - 128 & 4.19 & \textbf{20.72} & \textbf{0.0717} & \textbf{0.0103}\\
    \bottomrule
  \end{tabular}
  }
\end{table}

\subsubsection{Codebook Size}
The codebook size is a critical factor in the performance of VQ-based models, as it directly affects the expressiveness of the latent space and the quality of image reconstruction. However, a larger codebook size also requires significant system memory and computational resources, making it crucial to strike a balance between performance and efficiency. As a result, researchers often aim to optimize model performance while minimizing codebook size to reduce memory overhead and improve computational feasibility. In Table \ref{tab:codebook size}, we evaluate the performance of our Dual Codebook framework under different codebook sizes 1024 and 512. Our results indicate that increasing the codebook size enhances the model’s performance in FID score (3.24), demonstrating high-quality image reconstruction. However, even with a smaller codebook size of 512, our method still maintains strong performance, ensuring competitive image quality while significantly reducing memory usage. These findings highlight the effectiveness and adaptability of the Dual Codebook framework, as it leverages an optimized quantization strategy that remains robust across different codebook sizes. The ability to maintain competitive performance with a smaller codebook size underscores its efficiency and scalability, making it a practical choice for memory-constrained environments.

\begin{table}
  \centering
  \small
  \caption{Evaluation results of ablation study on MS-COCO dataset with codebook size of 1024 and 512. The best results are highlighted in bold.}
  \label{tab:codebook size}
  \resizebox{\columnwidth}{!}{ 
  \begin{tabular}{l ccccc}
    \toprule
    Models & Codebook Size  & FID$\downarrow$ & PSNR$\uparrow$ & $L_1\downarrow$ & $L_2\downarrow$ \\
    \midrule
    (i) Dual Codebook & 1024 & \textbf{3.24} & 20.48 & 0.0751 & 0.0108\\
    (ii) Dual Codebook & 512  & 4.19 & \textbf{20.72} & \textbf{0.0717} & \textbf{0.0103} \\
  
    \bottomrule
  \end{tabular}
  }
\end{table}


\subsubsection{Codebook Utilization}
We also examined the impact of our Dual Codebook method on codebook utilization using the MS-COCO dataset. As shown in Figure~\ref{fig:utility}, we compare our model against vanilla VQ-GAN \cite{esser2021taming} with deterministic quantization. The first row of this figure presents the codebook visualizations for both methods, revealing a significant difference in codebook updates between our approach and VQ-GAN. In this row, the blue grids represent intact and inactive code vectors, highlighting the codebook collapse issue in VQ-GAN. 

The second row illustrates the codebook usage for both models. The left blue graph (containing 1024 code vectors) depicts VQ-GAN’s codebook usage, further demonstrating the codebook collapse issue. In contrast, the red and blue graphs (each with 512 code vectors) correspond to the global and local codebooks in our proposed quantization method, respectively. This figure provides clear evidence of the Dual Codebook quantization method's effectiveness in enhancing codebook utilization and improving reconstructed image quality.

\section{Conclusion}
\label{sec:conclusion}
This paper presents a novel Dual Codebook framework for VQ-based image reconstruction models. The proposed Dual Codebook method is a simple yet effective approach that can be seamlessly integrated into standard VQ-GAN. It leverages two independent codebook optimization strategies, trained from scratch without relying on pretrained code vectors. Experimental results demonstrate that the Dual Codebook framework outperforms state-of-the-art VQ models in image modeling while maintaining the same codebook size.
\subsection{Limitations and Future Work}
As shown in our qualitative analyses (Figure~\ref{fig:qualitative-result} and in supplementary Figures~\ref{fig:_qualitative-result} and \ref{fig:_qualitative-result-ade}), the model sometimes fails to accurately reconstruct text and facial details in complex scenes as well as gradient color changes (Figure~\ref{fig:_qualitative-result-cub}), which are particularly challenging to capture. These challenging elements typically require precise high-frequency information that may not be optimally captured by our current architecture.
Future work could address these limitations through several avenues. First, exploring alternative transformer architectures for the global codebook could enhance the model's ability to capture complex gradients and high-frequency details. Second, incorporating additional perceptual losses specifically targeting facial features and text could improve reconstruction quality for these challenging elements while maintaining our lightweight codebook approach. These improvements would further strengthen the Dual Codebook framework while preserving its core advantages of training from scratch with compact codebook size.

\section{Acknowledgments}
This work was supported by the Cancer Prevention and Research Institute of Texas First Time Faculty Award (J.M.L.) and a University of Texas System Rising STARs Award (J.M.L.). We thank Jai Prakash Veerla, Biraaj Rout, and Allison Sullivan for their assistance.
 
{
    \small
    \bibliographystyle{ieeenat_fullname}
    \bibliography{main}
}


\clearpage
\onecolumn  
\setcounter{page}{1}

\section{Supplementary Data}
\setcounter{figure}{0}  
\renewcommand{\thefigure}{S\arabic{figure}}  

\setcounter{table}{0}  
\renewcommand{\thetable}{S\arabic{table}}  

While our Dual Codebook approach demonstrates strong performance across several datasets, we observed limitations in certain scenarios. Notably, we were unable to achieve state-of-the-art FID scores on the CUB-200~\cite{WahCUB_200_2011} dataset (Table~\ref{tab:cub200}), despite having lower $l_2$ losses. This suggests that our model struggles with capturing the gradient color changes in birds' feathers and background elements like water or sky. The discrepancy between perceptual metrics (FID) and pixel-wise losses ($l_1$ , $l_2$) indicates that our approach may prioritize local accuracy over global coherence in certain complex textures. As further experimentation is required to better understand these results, we have moved the detailed analysis on CUB-200 dataset to the supplementary material.

\begin{table}[h]
    \centering
    \renewcommand{\arraystretch}{1.2} 
    \caption{Performance comparison on CUB-200~\cite{WahCUB_200_2011} dataset.}
    \begin{tabular}{c c c c c} 
        \toprule
        \textbf{Models} & \textbf{FID$\downarrow$} & \textbf{PSNR$\uparrow$} & $\mathbf{L_1} \downarrow$ & $\mathbf{L_2} \downarrow$ \\
        \midrule
        VQ-VAE~\cite{van2017neural} & 54.92 & 24.38 & 0.0849 & 0.0183 \\
        VQ-GAN~\cite{esser2021taming} & 3.63  & 22.19 & 0.1051 & 0.0319 \\
        Gumbel-VQ~\cite{baevski2019vq} & 3.45  & 22.11 & 0.1048 & 0.0318 \\
        CVQ~\cite{zheng2023online} & 3.61  & 22.29 & 0.1034 & 0.0320 \\
        VQCT~\cite{zhang2024codebook} & \textbf{2.13}  & \textbf{25.35} & \textbf{0.0704} & 0.0160 \\
        \midrule
        Dual Codebook (Ours) & 5.93  & 20.8  & 0.0728 & \textbf{0.0102} \\
        \bottomrule
    \end{tabular}
    \label{tab:cub200}
\end{table}

\begin{figure}[h]
    \centering
    \includegraphics[width=\textwidth]{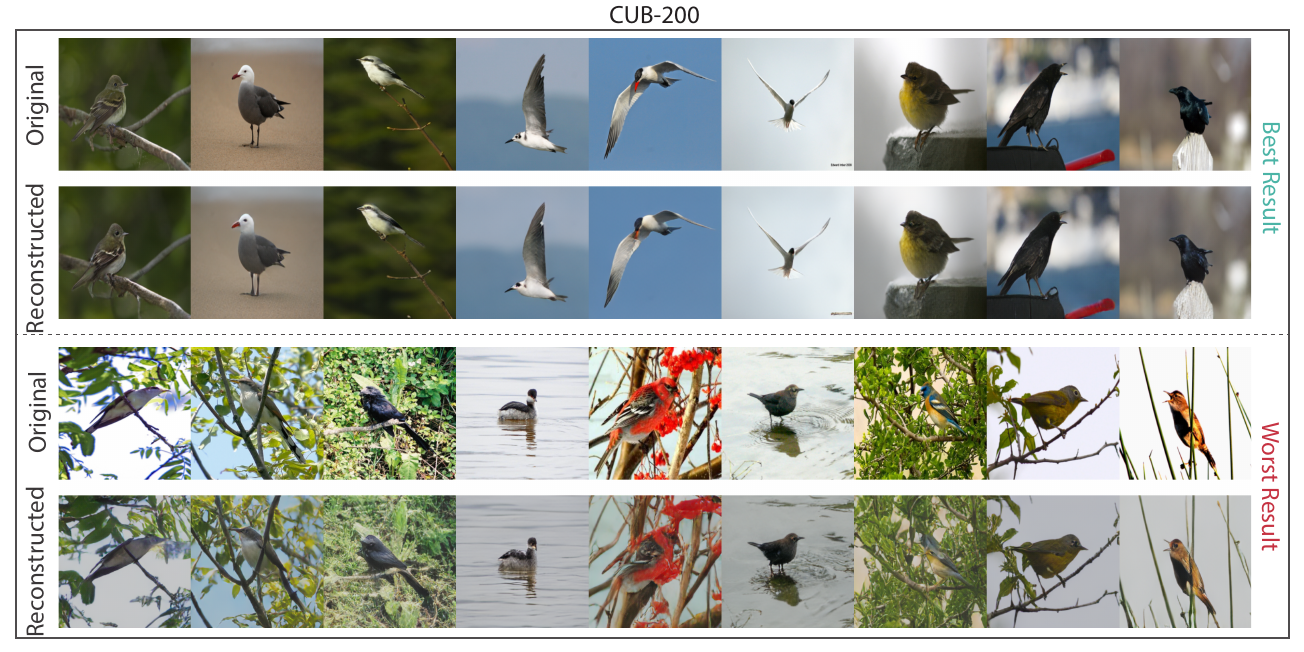}
    \caption{Best and worst reconstruction results on CUB-200 (test split) with Dual Codebook quantizer. The first and second rows show the best and worst cases based on PSNR value.}
    \label{fig:qualitative-result-cub}
\end{figure}

\begin{figure}[h]
    \centering
    \includegraphics[width=\textwidth]{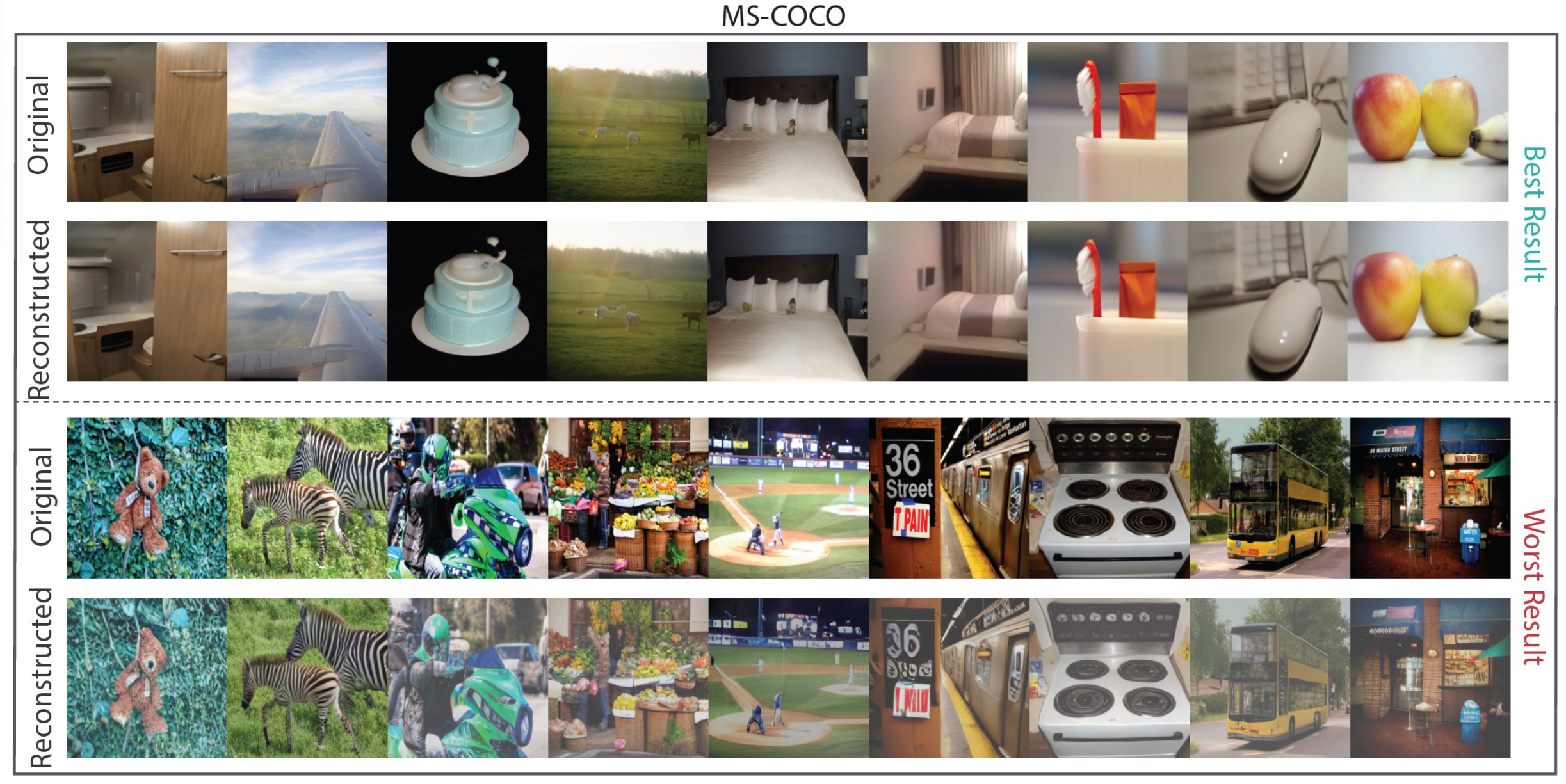}
    \caption{Best and worst reconstruction results on MS-COCO (test split) with Dual Codebook quantizer. The first and second rows show the best and worst cases based on PSNR value.}
    \label{fig:qualitative-result-coco}
\end{figure}

\begin{figure}[h]
    \centering
    \includegraphics[width=\textwidth]{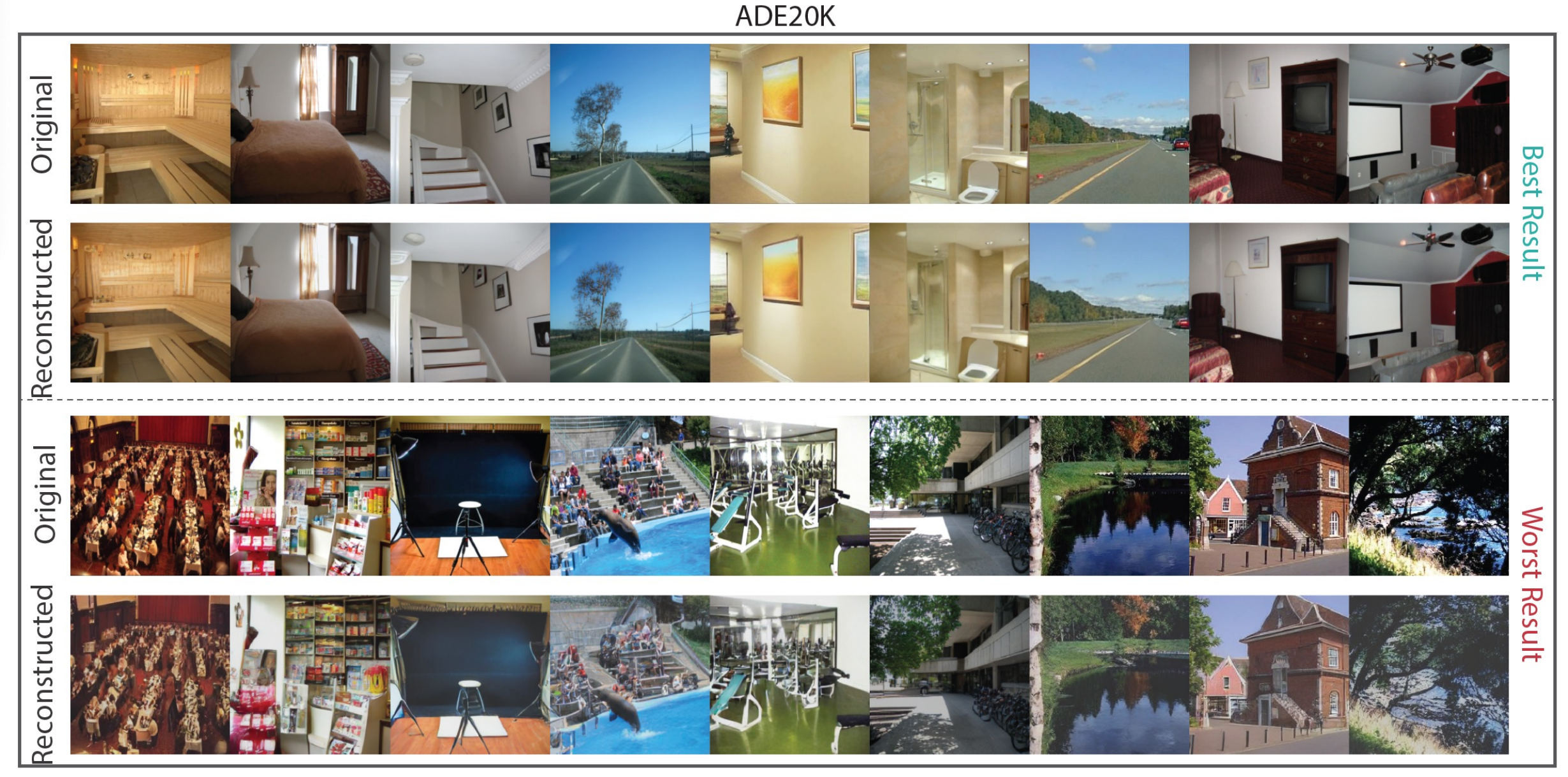}
    \caption{Best and worst reconstruction results on ADE-20K (test split) with Dual Codebook quantizer. The first and second rows show the best and worst cases based on PSNR value.}
    \label{fig:qualitative-result-ade}
\end{figure}

\begin{figure}[h]
    \centering
    \includegraphics[width=\textwidth]{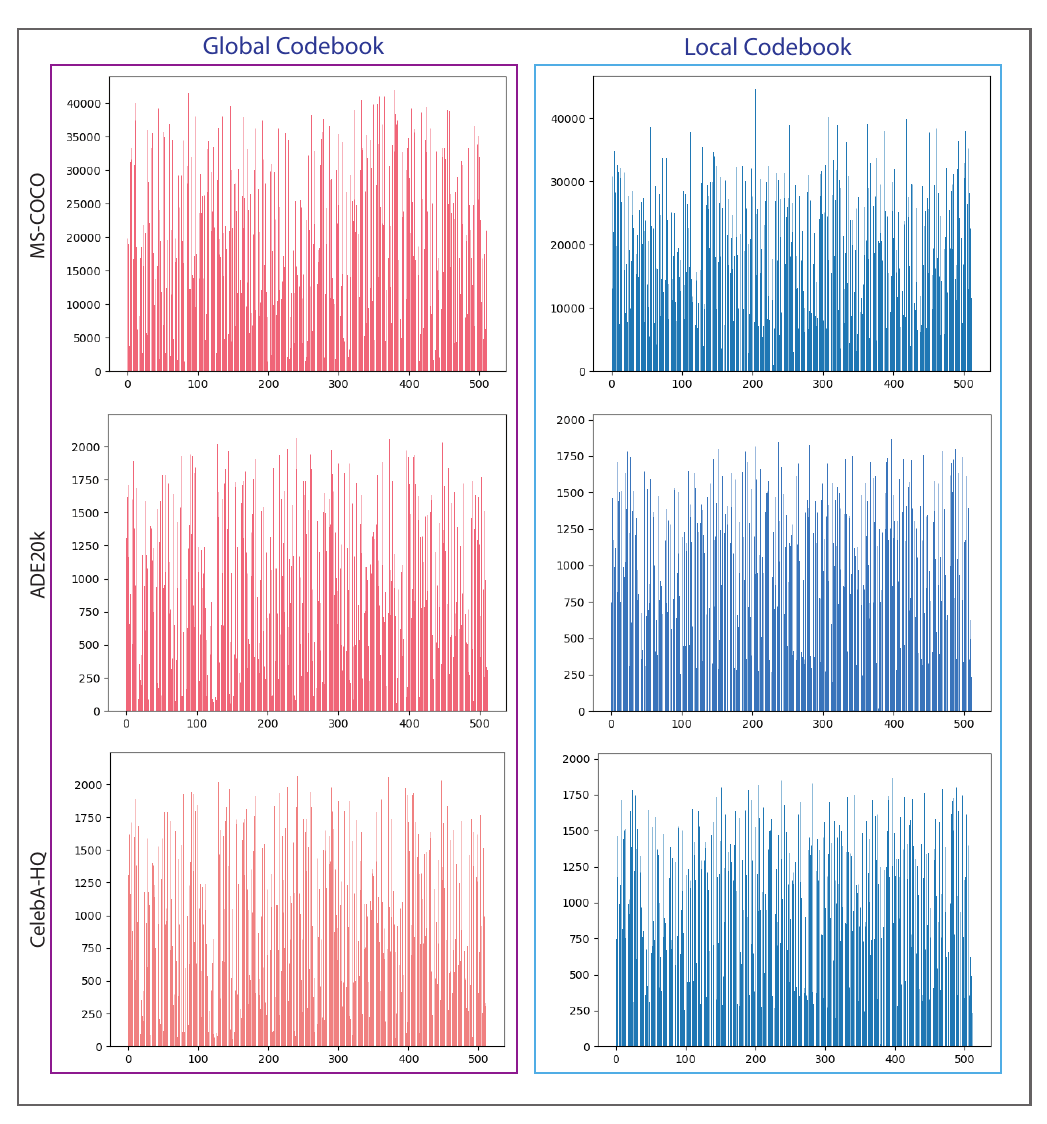}
    \caption{Utilization of global and local codebooks in the Dual Codebook method, shown in red and blue, respectively, across three datasets. This visualization demonstrates the effective use of both codebooks across all datasets, validating the method's robustness.}
    \label{fig:supp4-utility}
\end{figure}

\end{document}